\documentclass[sigconf]{acmart}

\usepackage{soul}
\usepackage{url}
\usepackage{hyperref}
\usepackage{inputenc}
\usepackage{graphicx}
\usepackage{amsmath,amsfonts}
\usepackage{booktabs}
\usepackage{multirow}
\usepackage{bm}
\usepackage{tikz}
\usepackage{subcaption}
\usepackage{algorithm}
\usepackage{placeins}
\usepackage{bm}
\usepackage{algpseudocode}
\newlength{\commentindent}
\usepackage{fancyhdr} 
\usepackage{lipsum}
\usepackage{float}

\DeclareMathOperator*{\argmin}{argmin}

\AtBeginDocument{%
  \providecommand\BibTeX{{%
    \normalfont B\kern-0.5em{\scshape i\kern-0.25em b}\kern-0.8em\TeX}}}

\copyrightyear{2022}
\acmYear{2022}
\setcopyright{acmcopyright}\acmConference[CIKM '22]{Proceedings of the 31st ACM International Conference on Information and Knowledge Management}{October 17--21, 2022}{Atlanta, GA, USA}
\acmBooktitle{Proceedings of the 31st ACM International Conference on Information and Knowledge Management (CIKM '22), October 17--21, 2022, Atlanta, GA, USA}
\acmPrice{15.00}
\acmDOI{10.1145/3511808.3557630}
\acmISBN{978-1-4503-9236-5/22/10}
\makeatletter
\def\algbackskip{\hskip-\ALG@thistlm}
\makeatother

\begin{document}


\title{Local Contrastive Feature Learning for Tabular Data}



\author{Zhabiz Gharibshah}
\affiliation{%
  \institution{Dept. of EECS, Florida Atlantic University}
  \city{Boca Raton, FL 33431}
  \country{USA}}
\email{zgharibshah2017@fau.edu}

\author{Xingquan Zhu}
\affiliation{%
  \institution{Dept. of EECS, Florida Atlantic University}
  \city{Boca Raton, FL 33431}
  \country{USA}}
\email{xzhu3@fau.edu}

\renewcommand{\shortauthors}{Zhabiz Gharibshah \& Xingquan Zhu}
\begin{abstract}
 Contrastive self-supervised learning has been successfully used in many domains, such as images, texts, graphs, etc., to learn features without requiring label information. 
 In this paper, we propose a new local contrastive feature learning (LoCL) framework, and our theme is to learn local patterns/features from tabular data. In order to create a niche for local learning, we use feature correlations to create a maximum-spanning tree, and break the tree into feature subsets, with strongly correlated features being assigned next to each other. Convolutional learning of the features is used to learn latent feature space, regulated by contrastive and reconstruction losses. Experiments on public tabular datasets show the effectiveness of the proposed method versus state-of-the-art baseline methods.
\end{abstract}

\ccsdesc[500]{Information systems~Data mining}


\keywords{Contrastive learning, self-supervised learning, tabular data}


\maketitle
\section{Introduction}
Tabular data, using rows (instances) and columns (features) to represent objects, are ubiquitous in nearly all applications \cite{gharibshah2020deep,autoint, anomaly2022}. Feature engineering is a traditional method to analyze the data and produce informative features for predictive modeling. 
Recently, self-supervised learning combined with deep learning methods to learn feature representations from unlabeled data has shown considerable success in different domains, especially for images, graphs and texts~\cite{SimCLR,bt,Zhu2020Graph,oord, contrastiveRanking}. Some studies have conducted to extend this success to tabular data where data samples are represented by vectors with different value types \cite{vime,subTab}. In practice, decision tree-based models like XGBoost are still known as strong non-gradient based models with a comparable or even superior performance \cite{2021contro,grinsztajn2022tree}. However, some advantages with deep learning methods like attention mechanism, pre-training parameters and providing an end-to-end data processing paradigm for training make them appealing for learning efficient feature representations \cite{tabnet2019, saint}.

Lack of clear feature relationships in tabular data, 
fully connected dense neural networks are typically used as a parametric method for training to consider the impact of all features on the target values in supervised setting \cite{revisiting2021,WellTunedMLP2021}. 
Some methods have been proposed to enable deep feature learning in contrastive learning paradigm for tabular data, however, they all use dense layer network 
~\cite{vime,scarf2021,ContrastiveMixup2021,anomaly2021}. 
The main drawback of dense layer is that they learn global patterns using all features. In many datasets (or learning tasks), patterns only involve a small number of features (not all features are useful). On the other hand, in real-world datasets, features are often subject to some correlations, which naturally results in local interactions \cite{han2019convolution}.
That motivates us to explore local pattern learning for tabular data. Here local learning is referring to that only a few number of features are involved in the pattern learning via the convolutional neural network (CNN) kernels. CNN networks have known as effective network design with parameter sharing to reduce model complexity 
to capture spatial connections between neighboring features with contiguous values. To address this problem in tabular data, we will create a niche of feature correlation by exploring a meaningful reordering of input features to apply convolutional kernels. 

To leverage convolutional feature learning, we develop a novel algorithm to use pairwise Pearson correlation coefficients between features as the metric to create a maximum spanning tree to connect all features followed by a depth-first-search traverse. It generates the new order of features being spatially correlated. In addition, we convert the definition of feature learning from a holistic format containing the whole feature values to multiple subsets of features created by feature splitting. We propose a self-supervised learning framework which leverages a 1-D convolutional denoising autoencoder \cite{DCAE2019} as a building block to capture correlations within a subset of the reordered input features and to maximize mutual information using contrastive comparisons between pair of subset embedding vectors.
We hypothesize that a deep neural network with convolutional operation on a local set of features, combined with contrastive and reconstruction optimizations in a self-supervised manner, can provide effective performance in classification downstream tasks.
\begin{figure*}[h]
    \begin{subfigure}{.5\textwidth}
  \begin{flushleft}
  \includegraphics[height=0.55\textwidth]{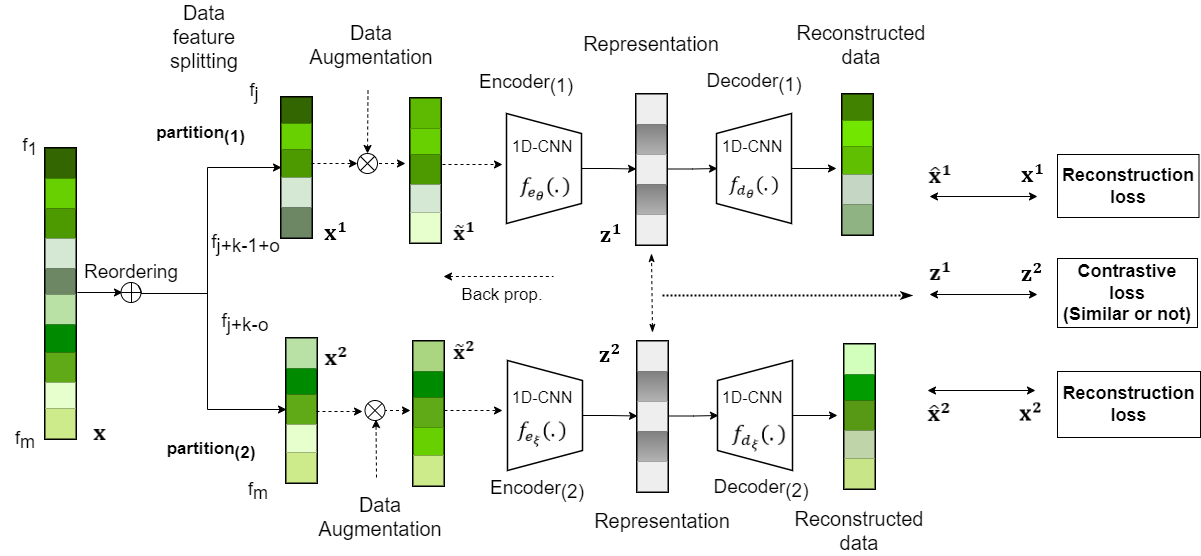}
    \caption{}
    \label{fig:proposed_model}
    \end{flushleft}    
    \end{subfigure}
    ~
    \begin{subfigure}{.5\textwidth}
  \begin{flushright}
  \includegraphics[height=0.4\textwidth]{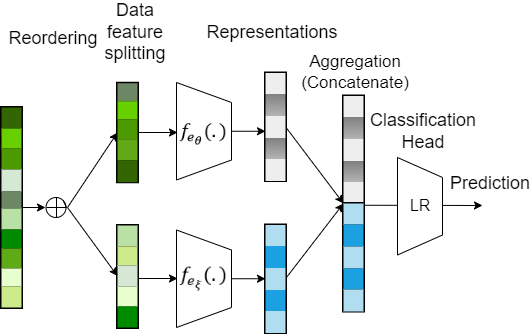}
    \caption{}
    \label{fig:fine_tune}
    \end{flushright}    
    \end{subfigure}    
    \vspace{-8mm}
    \caption{\small a)(\underline{Unsupervised pre-training}): workflow of the proposed method for self-supervised local contrastive learning: From left to right, the $m$ features of a tabular detest are partitioned into two subsets (with or without overlapping). A convolutional auto-encoder is trained from each subset, respectively. A contrastive loss is added to ensure latent features learned from feature partitions of the same instance are close to each other and to be distant if they are from two different instances; b)(\underline{Supervised learning}) Representation at the time for a prediction task consisting of training a linear classifier on top of frozen representations.}    
\end{figure*}

\begin{algorithm}
\footnotesize
\caption{\footnotesize LoCL: Local Contrastive Feature Learning}
\label{alg:alg1}
\begin{flushleft}
\hspace*{\algorithmicindent} \textbf{Inputs:} Augmentation $\mathcal{T}$, 
encoder ${f_e}_\theta$, encoder ${f_e}_\xi$, decoder ${f_d}_\theta$, decoder ${f_d}_\xi$, Batch size: $n$, encoder layer size: $d$, input feature indices: $\mathbb{F}$=[$f_1$, $f_2$, ..., $f_m$]
\end{flushleft}
\begin{algorithmic}[1]
\State \Comment{Reordering input features}
\State $\mathcal{M}\in \mathbb{R}^{m\times m}\leftarrow \text{Pearson}(\mathbf{X},\mathbb{F})$ \Comment {Calculate a pairwise feature correlation matrix}
\State $\text{MST} \leftarrow \text{Maximum Spanning Tree}(\mathcal{M})$ \Comment{Create a feature correlation maximum spanning tree considering input features $f_1$, $f_2$, ..., $f_m$ as nodes and top $m-1$ pearson correlations as the edges.}

\State $\mathbb{\bar{F}} \leftarrow \text{DFS}(MST)$ \Comment{Generate feature orders using the order of visited nodes by using a DFS traverse starting from a feature with the highest pairwise correlation.}
    
\State $[\mathbb{F}^1,\mathbb{F}^2] \leftarrow \text{split}(\bar{\mathbb{F}})$ \Comment{Divide the features into 2 subsets split uniformly from $\mathbb{\bar{F}}$} 
\For {sampled batch $\mathcal{B}:\{\textbf{X}|\{{\textbf{\text{x}}}\}^n_{k=1}\}$}
    \ForAll{$k$=1 to n}   
    \State \Comment{Apply augmentations and get network outputs}       
    \State $\textbf{x}^1, \textbf{x}^2 = \textbf{x}[k,\mathbb{F}^1], \textbf{x}[k,\mathbb{F}^2]$
    \State $\textbf{\text{z}}^1, \textbf{\text{z}}^2 =  {f_{e_1}}(\mathcal{T}_{1}(\textbf{x}^1)), {f_{e_2}}(\mathcal{T}_{2}(\textbf{x}^2))$
    \State $\hat{\textbf{\text{x}}}^{1}, \hat{\textbf{\text{x}}}^{2}$ = ${f_{d_1}}(\textbf{\text{z}}^1), {f_{d_2}}(\textbf{\text{z}}^2)$
    
    \State \Comment{reconstruction and contrastive loss}
    \State ${\mathcal{L}_{r_1}}[k],{\mathcal{L}_{r_2}}[k] = \lVert  \hat{\textbf{\text{x}}}^2-\textbf{\text{x}}^2 \rVert^2_2, \lVert  \hat{\textbf{\text{x}}}^1-\textbf{\text{x}}^1 \rVert^2_2$ 
    
 

 
\State ${\mathcal{L}_{r}}[k] = \dfrac{1}{2}\sum_{i}^{2}({\mathcal{L}_{r_i}}[k])$
        \State ${\mathcal{L}_{c}}[k] =  l_{c}(\textbf{z}^j,\textbf{z}^l)[k]$ 
    \State $\mathcal{L}[k] = {\mathcal{L}_{c}}[k]+ \alpha~{\mathcal{L}_{r}}[k]$

    \EndFor
    
    \State $\nabla_{\theta,\xi} \mathcal{L}$ \Comment{Calculate gradients and update all trainable parameters}
\EndFor

\end{algorithmic}
\begin{flushleft}
\hspace*{\algorithmicindent} \textbf{Output:} encoder ${f_e}_\theta$,${f_e}_\xi$        
\end{flushleft}

\end{algorithm}
\section{Problem Definition}
A tabular dataset $\mathcal{D}=\{\textbf{x}_i\}_1^N$ consists of $N$ instances and $m$ features as m-dimensional vector $\textbf{x}_i \subset \mathcal{X} \in \mathbb{R}^m$ among which a small subset of data samples is labeled, i.e. $\mathcal{D}_L = \{\textbf{x}_i,y_i\}_1^N$ where  $\mathcal{D}_L \subset \mathcal{D}$, $|\mathcal{D}|\gg |\mathcal{D}_L|$ and $y_i \subset \mathcal{Y} \in \mathbb{R}$ is a discrete label set containing two or more categorical values. In a supervised setting, learning a predictive model $f: \mathcal{X} \rightarrow \mathcal{Y}$ is optimized by using a supervised loss function (e.g. cross-entropy loss function). But when a small set of labeled data is available it may lead to overfitting. Therefore, we develop an unsupervised representation learning to use unlabeled data to handle this problem 
to learn a feature mapping function $f(.):\mathcal{X}\rightarrow \mathcal{Z}$ where $\textbf{z}= f(\textbf{x})$ is a feature representation of input sample. In self-supervised learning, in absence of the label information, the representation $\textbf{z}$ is optimized using a self-supervised loss function according to pre-defined pseudo labels. It is based on a pre-defined notion of similarity (positive labels) between embedding vectors of two pairs of data points versus the pre-defined dis-similarity(negative labels) between other pairs of data points.

For a given dataset, we use a self-supervised learning framework (as shown in Figure~\ref{fig:proposed_model}) to learn latent feature representations. During this state, no label information is available for model learning (\textit{i.e.} a pure self-supervised learning fashion). In order to validate the quality of the latent features for a classification task, during the fine-tuning stage, as shown in Figure~\ref{fig:fine_tune}, we use a small number of labeled instances to train a classification model and validate the performance of the classifier trained using representations made from the concatenation of embedding vectors of feature subsets.
\begin{figure}[h]
   \centering
   \includegraphics[width=0.45\textwidth]{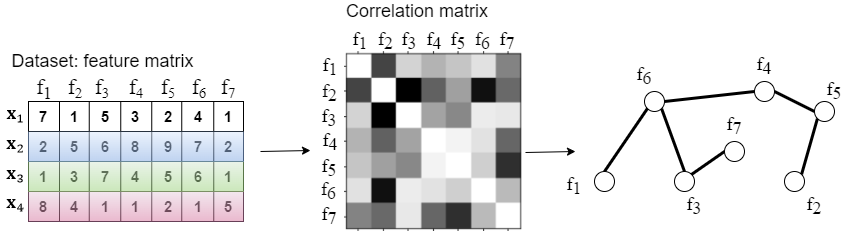}
   \vspace{-3mm}
    \caption{\small A conceptual view of feature ordering using maximum spanning tree. Left: A toy tabular dataset containing four instances and seven features; Middle: Correlation between features (Lighter color denotes stronger correlation); Right: maximum spanning tree constructed using feature correlation.}
    \label{fig:feature ordering}
\end{figure}
\section{Proposed Framework}


Self-supervised learning is 
typically defined through pretext tasks after applying data augmentation on data. Rather than doing regular contrastive learning on the entire feature set, in the proposed method, we consider a localized version of representation learning when the input features are split into different subsets. 
This method is developed based on the idea that in regular holistic unsupervised feature learning methods like Denoising Autoencoders\cite{dae}, treating all input features equally may not be effective since it assumes that either they are independent or they contribute with almost similar levels to represent data. Pixel values in images have spatial or sequential correlations, making image cropping as one of the common operations for in data augmentation tasks \cite{BYOL,subTab}. This motivates us to propose a method for tabular domain to divide the single neural network encoder into multiple modules being responsible to learn representation for the subsets of input features in the representation learning process. 

Figure \ref{fig:proposed_model} illustrates the model using two subsets of features. The model contains two main components: (a) feature reordering and partitioning, which reorganizes features into subsets with reordered adjacency relationships. It is assumed that the model typically has non-overlapping feature subsets but some level of overlaps between feature subsets are also allowed; and (b) local contrastive learning which learns representations by combining convolutional autoencoder and contrastive optimization.

\subsection{Tabular Feature Reordering}
Since the order of features in tabular data does not necessarily follow spatial correlation like neighboring pixels in images or sub-sequential frames of videos, it is preceded by our own feature reordering step. This step leverages mutual correlation between input features to reorder them and let a specific deep learning structure like a convolutional neural layer in the encoder and decoder components be applied to learn effective local feature representations. Without this, although applying convolutional deep learning models is doable, but they do not come with any intuition. We propose to use mutual feature correlations to determine the new order of input features. In Fig. \ref{fig:feature ordering}, we use Pearson correlation coefficients as the metric and create a feature-feature correlation matrix $\mathcal{M} \in \mathbb{R}^{m\times m}$ from all mutual Pearson correlation values. Assuming all features as nodes of a fully-connected graph in which the weights of the edges between nodes are presented by 
the absolute value of correlations, we create a maximum spanning tree 
where the sum of all weights of connecting edges is the maximum value as possible. The order of features is determined by a depth-first search starting from nodes that the edge with the highest value connects in the tree.

\subsection{Local Contrastive Learning}
After calculating the order of input features, we convert each instance into a data sample with the new order of features where feature values in the adjacent one-dimensional window of features contain Pearson correlations being similar to spatial correlation in images. So we use 1-D CNN in the deep learning network to learn new feature representations. Each sample in input data is divided by two feature subsets. For self-supervised learning, we apply a stochastic data augmentation on sample subsets through masking which randomly generates a binary matrix with a batch of data related to input feature subsets. Each vector in the binary mask is randomly sampled from a Bernoulli distribution with a pre-defined probability parameter. The corrupted version of each sample in either batch of feature subsets is calculated as follows:
\begin{equation}
\label{obj_function}
\widetilde{\bm{\mathrm{{x}}}}_i = \bm{\mathrm{{x}}}_i\odot (\mathbf{1}-\textbf{\text{m}}) + \bm{\mathrm{\bar{x}_i}}\odot \textbf{\text{m}}
\end{equation}
It is then fed to a designated encoder to transform the input data into a representation with respect to the selected subset, then the corresponding decoder is responsible to recover the original input data. Given the introduced data augmentation procedure, the optimization is done by using the linear combination of two loss functions with respect to the input feature value reconstruction task and the contrastive representation estimation task as follows:
\begin{equation}
\small
\label{final_obj_function}
\begin{aligned}
\argmin_{\theta,\xi} \mathcal{L}_c + \alpha~\mathcal{L}_r
\end{aligned}
\end{equation}
where $\theta$ and $\xi$ refer to trainable weights in encoder and decoder networks according to a pair of feature subsets. $\mathcal{L}_c$ is our contrastive loss function and $\mathcal{L}_r$ is the reconstruction loss function. We use the following loss function for a de-noising task to predict the original feature values from corrupted data vectors:
\begin{equation}
\small
\label{r_obj_function}
\begin{aligned}
\mathcal{L}_{r}(\hat{\textbf{\text{X}}},\textbf{\text{X}})= \dfrac{1}{2}\sum_{j}^{2}\dfrac{1}{n}\sum_{i}^{n} (f_{d_j}(\textbf{\text{Z}}^j)-\textbf{\text{X}}^j)^2
\end{aligned}
\end{equation}

The reconstruction loss value is calculated as the average of minimum squared error between the reconstructed vectors and the original input over two feature subsets. The contrastive loss function between pair of representations of feature subsets are calculated via Barlow-twins contrastive loss  function \cite{bt} by calculating a cross-multiplication matrix $\textbf{\text{C}} = {\textbf{Z}^1}^T\cdot {\textbf{Z}^2}^T \in \mathbb{R}^{d \times d}$ where $d$ is the the dimension of hidden layer in the model. It is computed through the dot-product of the batch of normalized embedding vectors. The matrix is going to be optimized to be equal to the identity matrix $I_p$ in the following contrastive loss function:
\begin{equation}
\small
\label{bt_loss_func}
\mathcal{L}_{c}(\textbf{\text{Z}}^1,\textbf{\text{Z}}^2) = |\textbf{\text{C}}-\textbf{\text{I}}_p|^2 = \sum\limits_{i}(1-\textbf{\text{C}}{[i,i]})^2+ \lambda\sum\limits_{i}{\sum\limits_{j\neq i}(\textbf{\text{C}}{[i,j]})^2}
\end{equation}

The contrastive loss function encourages the similarity of the pair of input feature subsets that are split from the same batch of data. The model learns from the level of inconsistency between a pair of corrupted data and also from the correlation between the non-mask area in pair of feature subsets. We expect that the aggregated representations learned by the encoder components can be employed in the fine-tuning step to be used in classification tasks. Algorithm \ref{alg:alg1} shows the pseudo-code of the proposed method.

\section{Experiments}
\subsection{Experimental Settings}
\paragraph{\textbf{Datasets}} We evaluate the performance of the proposed method on six 
benchmark datasets publicly available on the UCI repository \cite{Dua:2019}. Table \ref{tbl:db_list} describes the statistics of the datasets. 
\vspace{-.1cm}
\begin{table}[h]
 \centering
\footnotesize
 \caption{\footnotesize{Basic statistics of benchmark datasets used in the experiments}} 
\vspace{-4mm} 
 \label{tbl:db_list}
 \begin{tabular}{l|l|l|l}
  \toprule[1pt]
Dataset         & \# of features  & \# of Samples  & \# of Classes \\\hline
MNIST           & 784      & 70,000  & 10      \\
Income & 14    & 48,842     & 2       \\
BlogFeedback & 280    & 60,021     & 2       \\
Diabetic Retinopathy         & 20      & 1151  & 2      \\
Wall-following & 55    & 5456     & 4       \\
Gas sensor array drift & 128    & 13,910     & 4       \\
 \bottomrule[1pt]
 \end{tabular}
 \end{table}

\begin{table*}[!htp]
\setlength{\tabcolsep}{3pt} 
\small
\caption{\small{Target prediction results; Comparison between LoCL and the baseline
methods. The evaluation metrics are mean $\pm$ std. of accuracy  scores over 5-fold cross validation for the classification task. The number of latent dimension is shown within parentheses}}
\vspace{-4mm} 
\label{tbl:main_results}
\begin{tabular}{l|l|l|l|l|l|l|l|c}
\toprule[1pt]
\multicolumn{2}{c|}{\multirow{1}{.1cm}{\textbf{Model/Dataset}}}      & MNIST(256)                       & INCOME\small{(512)}                 & BLOG\small{(1024)}& Diabetes\small{(64)}& Wall-follow.\small{(64)}  & Gas sensor\small{(512)}& \small{Average}                \\\hline
\multirow{4}{1.4cm}{\textbf{Supervised\\Learning}} & LR            & 0.9221 $\pm$ 0.001 & 0.8243 $\pm$ 0.003 & 0.7728 $\pm$ 0.003 & 0.7280 $\pm$ 0.025 & 0.7008 $\pm$ 0.020 & 0.9902 $\pm$ 0.002 & 0.823 $\pm$ 0.01 \\
                                         & MLP           & 0.9743 $\pm$ 0.001 & 0.8501 $\pm$ 0.003 & 0.7885 $\pm$ 0.004 & 0.6977 $\pm$ 0.019 & 0.9129 $\pm$ 0.007 & 0.9891 $\pm$ 0.006 & 0.869 $\pm$ 0.01 \\
                                         & RF            & 0.9664 $\pm$ 0.002 & 0.8571 $\pm$ 0.003 & 0.8272 $\pm$ 0.003 & 0.6681 $\pm$ 0.035 & 0.9940 $\pm$ 0.003 & 0.9942 $\pm$ 0.002 & 0.885 $\pm$ 0.01 \\
                                         & XGBoost       & 0.9041 $\pm$ 0.002 & 0.8555 $\pm$ 0.004 & 0.8249 $\pm$ 0.003 & 0.6951 $\pm$ 0.042 & 0.9962 $\pm$ 0.002 & 0.9729 $\pm$ 0.002 & 0.875 $\pm$ 0.01 \\\hline
\multirow{5}{1.4cm}{\textbf{Self-supervised\\Learning}}                 & DAE~\cite{dae} & 0.8982 $\pm$ 0.006 & 0.8222 $\pm$ 0.003 & 0.7201 $\pm$ 0.002 & 0.6273 $\pm$ 0.031 & 0.6642 $\pm$ 0.025 & 0.9448 $\pm$ 0.009 & 0.779 $\pm$ 0.01 \\
& Conv-DAE & 0.9518 $\pm$ 0.003 & 0.8324 $\pm$ 0.005 & 0.7406 $\pm$ 0.006 & 0.5777 $\pm$ 0.061 & 0.6557 $\pm$ 0.021 & 0.9692 $\pm$ 0.003 & 0.788 $\pm$ 0.02  \\
& Barlow-twins~\cite{bt} & 0.9431 $\pm$ 0.001 & 0.8378 $\pm$ 0.004 & 0.7507 $\pm$ 0.003 & 0.6386 $\pm$ 0.032 & 0.7269 $\pm$ 0.027 & 0.9807 $\pm$ 0.003 & 0.813 $\pm$ 0.01  \\
& SimCLR~\cite{SimCLR} & 0.9432 $\pm$ 0.003 & 0.8434 $\pm$ 0.005 & 0.7569 $\pm$ 0.003 & 0.6238 $\pm$ 0.055 & 0.6946 $\pm$ 0.022 & 0.9724 $\pm$ 0.006 & 0.806 $\pm$ 0.02  \\
                                         & VIME~\cite{vime}          & 0.9377 $\pm$ 0.002 & 0.8458 $\pm$ 0.004 & 0.7406 $\pm$ 0.004 & 0.6186 $\pm$ 0.057 & 0.7382 $\pm$ 0.029 & 0.9628 $\pm$ 0.006 & 0.807 $\pm$ 0.02 \\\cline{2-9}
                                         & LoCL  & \textbf{0.9540} $\pm$ \textbf{0.002} & \textbf{0.8461} $\pm$ \textbf{0.005} & \textbf{0.7783} $\pm$ \textbf{0.004} & \textbf{0.6438} $\pm$ \textbf{0.037} & \textbf{0.7479} $\pm$ \textbf{0.011} & \textbf{0.9825} $\pm$ \textbf{0.004} & \textbf{0.825} $\pm$ \textbf{0.01}\\
\bottomrule[1pt]                                         
\end{tabular}
\end{table*}
\begin{table}[!htbp]
\small
   \caption{\small{Ablation studies to compare to the impact different encoder and the ordering of input features; average scores and standard deviation over all datasets are reported based on 5-fold cross-validation}} 
\vspace{-4mm}    
 \label{tbl:ablation study}
\begin{tabular}{l|l|l}
\toprule[1pt]
Model Variants           & Accuracy & Std    \\\hline
LoCL              & \textbf{0.8254}   & 0.01 \\
LoCL - Dense layer       & 0.8123   & 0.02 \\
LoCL - Random ordering   & 0.8013   & 0.01 \\
LoCL - Original order    & 0.8251   & 0.01 \\
LoCL - Interleaved order & 0.8070   & 0.01\\
 \bottomrule[1pt]
\end{tabular}

\end{table}
\vspace{-.3cm}
\paragraph{\textbf{Implementation details}}In all experiments, we first use data pre-processing techniques to transform raw data into well-formed data formats.  
For the image dataset like MNIST, the pixel values follow approximately a Gaussian distribution. To normalize data before running experiments, we apply a simple min-max normalization to put input values in the range of [0,1]. 
For the other datasets, we apply standardization to get z-normalized data. The categorical features in the datasets like the adult income dataset are one-hot-encoded. Furthermore, we also assume that all features have non-zero standard deviation. Otherwise, we discard them for training procedures. We evaluate the performance of all studied models through 5-fold stratified cross-validation in which 90\% of samples in the training data are randomly used as un-labeled data for the pre-training step. In the experiments setup, we use three 1-D CNN layers followed by max-pooling and up-sampling layers and LeakyReLU activations in the body of the encoder and the decoder components respectively. We use RMSProp optimizer with a learning rate of 0.001. We set Bernoulli probability parameter to 0.3 in the data augmentation step. The optimal value of hyper-parameters in the model (like trading parameter $\alpha$, the kernel size in convolutional networks, \textit{etc.}) are selected via cross-validation. All models are trained for a maximum of 200 epochs with an early stopping mechanism. After training the model, the trained model are used to transfer the remaining 10\% of training labeled data to the new feature space, and train classifier. 
\paragraph{\textbf{Baselines}}We compare LoCL with the following baseline:
\begin{description}
  \item[$\bullet$ DAE~\cite{dae}:] a denoising autoencoder augmented with multiplicative mask-out noise.
  \item[$\bullet$ Conv-DAE:] a 1-D denoising convolutional autoencoder
  \item[$\bullet$ Barlow-Twins~\cite{bt}:] a contrastive learning model with MLP as the encoder to do invariance optimization. 
  \item[$\bullet$ SimCLR~\cite{SimCLR}:] a contrastive learning model with MLP as the encoder to maximize mutual info using InfoNCE optimization. The projector in the model is skipped.
  \item[$\bullet$ VIME~\cite{vime}:] a model, which attachs a mask estimating decoder and a feature estimating decoder on top of the encoder.
\end{description}
\subsection{Experimental Results}
Table \ref{tbl:main_results} demonstrates the performance of different supervised and self-supervised methods. We assess the performance of different baseline methods using accuracy metric on the separate testing data. For comparison purposes, we include supervised learning results which use label information of all training data to train classifiers (LR, MLP, RF, \& XGBoost). This demonstrates the upper bound of self-supervised learning (which only uses 10\% of label information to train the classifier). 

According to the results, we can see in holistic autoencoder models, convolutional-based DAE models could obtain better performance than simple DAEs. Comparing with previous contrastive learning models like SimCLR and Barlow-Twins VIME models, we see improvements in accuracy scores in some datasets. It shows the importance of the self-supervised paradigm to learn informative representations for the downstream task. As for LoCL, it combines reconstruction and contrastive representation learning in one paradigm through local feature learning which leads to a superior performance against the state-of-the-art self-supervised methods.
\paragraph{Ablation Study}
We have conducted additional ablation studies to measure separately the impacts of two main components of the proposed method LoCL including the feature order and local feature learning through convolutional encoding and decoding on the classification performance. We create variants of the proposed method when we vary the structure of the model from convolutional neural network to a network with dense layers. We also investigate the effect of different feature orders in the other variants of the model when we use random ordering, original feature order, and interleaved (every other) feature order. Table \ref{tbl:ablation study} confirms that the proposed feature ordering approach, along with local self-supervised learning empowered with convolutional networks, makes a great improvement in the performance. 
\section{Conclusion}
In this paper, we introduced a new self-supervised method for learning feature representations for tabular data. 
We argued that existing methods largely rely on dense networks to learn feature representation, where dense networks aim to learn global patterns from all features. Since not all features are useful for learning tasks, and features often impose interactions, it is, therefore, more effective to learn local features. Alternatively, our model proposes a new feature reordering using feature-feature correlations and applies local feature learning to reordered feature subsets by using convolutional neural network modeling, combined with contrastive self-supervised learning. Experiments confirm the performance gain of the proposed method versus the state-of-the-art methods. 

\begin{acks}
This research is supported by the US National Science Foundation (NSF) through Grants IIS-1763452 and IIS-2027339. 
\end{acks}
\bibliographystyle{ACM-Reference-Format}
\bibliography{references}










\end{document}